\documentclass{article}
\usepackage{amsmath}
\usepackage{arxiv}
\usepackage[utf8]{inputenc} 
\usepackage[T1]{fontenc}    
\usepackage{hyperref}       
\usepackage{url}            
\usepackage{booktabs}       
\usepackage{amsfonts}       
\usepackage{nicefrac}       
\usepackage{microtype}      
\usepackage{graphicx}
\graphicspath{ {./images/} }

\title{MIA-Mind: A Multidimensional Interactive Attention Mechanism Based on MindSpore}

\author{
  Zhenkai Qin$^{1,2,3}$ \thanks{These authors contributed equally to this work.} \\
  $^{1}$School of Information Technology \\
  $^{2}$Network Security Research Center \\
  $^{3}$Big Data and Policing Technology Laboratory\\
  Guangxi Police College\\
  Nanning, Guangxi, China \\
  \texttt{qinzhenkai@gxjcxy.edu.cn} \\
  \And
  Jiaquan Liang$^{2}$ \thanks{These authors contributed equally to this work.}\\
  School of Information Technology \\
  Guangxi Police College \\
  Nanning, Guangxi, China \\
  \texttt{liangjiaquan@gxjcxy.edu.cn} \\
  \And
  Qiao Fang$^{2}$ \thanks{These authors contributed equally to this work.} \\
  School of Information Technology \\
  Guangxi Police College \\
  Nanning, Guangxi, China \\
  \texttt{fangqiao@gxjcxy.edu.cn} \\
}

\begin{document}
\maketitle

\begin{abstract}
Attention mechanisms have significantly advanced deep learning by enhancing feature representation through selective focus. However, existing approaches often independently model channel importance and spatial saliency, overlooking their inherent interdependence and limiting their effectiveness. To address this limitation, we propose MIA-Mind, a lightweight and modular Multidimensional Interactive Attention Mechanism, built upon the MindSpore framework. MIA-Mind jointly models spatial and channel features through a unified cross-attentive fusion strategy, enabling fine-grained feature recalibration with minimal computational overhead. Extensive experiments are conducted on three representative datasets: on CIFAR-10, MIA-Mind achieves an accuracy of 82.9\%; on ISBI2012, it achieves an accuracy of 78.7\%; and on CIC-IDS2017, it achieves an accuracy of 91.9\%. These results validate the versatility, lightweight design, and generalization ability of MIA-Mind across heterogeneous tasks. Future work will explore the extension of MIA-Mind to large-scale datasets, the development of adaptive attention fusion strategies, and distributed deployment to further enhance scalability and robustness.
\end{abstract}

\keywords{MindSpore Framework \and Multidimensional Attention \and Feature Recalibration \and Lightweight Neural Networks}

\section{Introduction}

Attention mechanisms have become a cornerstone in deep learning, enabling neural networks to dynamically prioritize informative features while suppressing less relevant information. Their integration has led to significant performance improvements across various domains, including image recognition, object detection, and natural language processing. Despite these successes, most existing attention mechanisms independently model either spatial saliency or channel importance, overlooking the intricate interdependencies between these feature dimensions. This separation often limits the expressiveness of feature representations, particularly in complex tasks requiring holistic context modeling. Overcoming this limitation necessitates the development of strategies that can jointly capture multidimensional feature relationships while maintaining computational efficiency.

Recent research efforts have sought to address the limitations of independent attention modeling by developing multidimensional attention mechanisms that jointly consider spatial and channel-wise features. These methods aim to enhance feature representation by capturing richer cross-dimensional dependencies. However, many existing approaches still rely on simple concatenation or additive fusion strategies, which inadequately model the complex interactions between spatial and channel contexts, resulting in suboptimal feature recalibration. Furthermore, some multidimensional attention designs introduce significant computational overhead due to the use of multiple attention branches, high-dimensional transformations, or global self-attention operations across the full feature map. This increased complexity not only impacts inference efficiency but also limits their scalability and applicability in real-world scenarios where resource constraints and deployment requirements are critical considerations.

In response to these challenges, we propose \textbf{MIA-Mind}, a lightweight and modular \textit{Multidimensional Interactive Attention Mechanism} implemented within the MindSpore framework. MIA-Mind introduces a unified cross-attentive fusion strategy that simultaneously models spatial and channel dependencies, enabling fine-grained feature recalibration with minimal additional computational overhead. The architecture is composed of three key modules: a global feature extractor that separately encodes spatial and channel information, an interactive attention generator that dynamically fuses these features via cross-multiplicative weighting, and a dynamic reweighting module that adaptively enhances salient feature responses.

To validate the effectiveness and generalizability of MIA-Mind, we conduct extensive experiments across three representative tasks: (i) image classification using ResNet-50~\cite{he2016deep} on the CIFAR-10 dataset; (ii) medical image segmentation using U-Net~\cite{ronneberger2015u} on the ISBI2012 dataset~\cite{ronneberger2015u}; and (iii) network traffic anomaly detection using a CNN-based architecture on the CIC-IDS2017 dataset~\cite{sharafaldin2018toward}. Experimental results demonstrate that MIA-Mind consistently enhances accuracy, improves boundary delineation, and increases anomaly detection sensitivity across all evaluated domains.

The main contributions of this work are summarized as follows:
\begin{itemize}
    \item We identify and address the limitation of existing attention mechanisms in jointly modeling spatial and channel dependencies.
    \item We design MIA-Mind, a novel multidimensional interactive attention module that enables efficient and dynamic feature recalibration within a lightweight framework.
    \item We validate the versatility and practicality of MIA-Mind across classification, segmentation, and anomaly detection tasks, demonstrating its effectiveness within the MindSpore ecosystem.
\end{itemize}

\section{Related Work}

\subsection{Evolution of Attention Mechanisms in Deep Learning}

Attention mechanisms have emerged as a fundamental component in enhancing the representational capacity of deep neural networks by enabling dynamic feature recalibration. Early works such as the Squeeze-and-Excitation (SE) network~\cite{hu2018squeeze} pioneered channel-wise attention by adaptively reweighting feature channels based on global context information. Building upon this concept, CBAM~\cite{woo2018cbam} introduced a sequential application of channel and spatial attention modules, albeit treating them independently. Subsequent advancements, including ECA-Net~\cite{wang2020eca} and BAM~\cite{park2018bam}, sought to improve computational efficiency and receptive field adaptation. ECA-Net utilized efficient local cross-channel interaction to reduce overhead, while BAM employed bottleneck structures to enhance attention modeling. Nevertheless, these designs largely maintained separable attention pathways, limiting their ability to capture the intrinsic interdependencies between spatial and channel features. More recently, Transformer-based architectures~\cite{dosovitskiy2021image} leveraged global self-attention to model long-range dependencies across input elements. Despite offering strong global reasoning capabilities, such models often suffer from high computational complexity, posing challenges for deployment in resource-constrained environments. Overall, the problem of jointly modeling spatial and channel interactions within a lightweight framework remains insufficiently addressed.

\subsection{Multidimensional Attention Mechanisms: Progress and Limitations}

In response to the limitations of independently modeling attention dimensions, several multidimensional attention mechanisms have been proposed. The Dual Attention Network (DANet)~\cite{fu2019dual} introduced parallel spatial and channel attention branches to capture semantic dependencies for scene segmentation. However, its additive combination strategy limited the expressive synergy between different dimensions. Triplet Attention~\cite{misra2021rotate} explored rotated triplet branch structures to align and apply attention across three orientations, yet incurred considerable computational costs due to the use of multiple attention pathways.Coordinate Attention (CA)~\cite{hou2021coordinate} enhanced localization by encoding positional information into channel attention maps through factorized convolution operations. While effective for mobile-friendly designs, CA may compromise fine-grained spatial modeling. More recently, axial attention~\cite{wang2020axial} has been proposed to decompose global self-attention into two one-dimensional attentions along height and width axes, effectively reducing computational complexity. However, axial attention treats each dimension separately, thereby limiting the modeling of intricate cross-dimensional interactions.Although these methods have advanced the modeling of multidimensional dependencies, they either suffer from incomplete cross-dimensional integration or impose non-negligible computational burdens. These observations motivate the development of a more unified and lightweight attention mechanism capable of efficiently capturing joint spatial-channel feature relationships, as proposed in this work.

\subsection{The MindSpore Framework: Technical Advantages for Modular Attention Design}

MindSpore is an open-source deep learning framework developed by Huawei, designed to facilitate efficient training and deployment across cloud, edge, and device platforms. It supports both dynamic and static computation graph modes, enabling flexible control flow while maintaining strong optimization capabilities. MindSpore incorporates advanced graph optimization strategies, including operator fusion and memory-efficient scheduling, which significantly reduce computational latency and memory footprint. These features are particularly beneficial for lightweight models with intensive attention operations. Additionally, the modular Cell abstraction system in MindSpore allows seamless encapsulation of complex modules, enabling flexible integration of customized mechanisms such as MIA-Mind into various backbone architectures. This modularity and efficiency make MindSpore a favorable platform for deploying the proposed attention model in diverse real-world scenarios.

\subsection{MIA-Mind: A Multidimensional Interactive Attention Mechanism Based on MindSpore}

In this work, we propose \textbf{MIA-Mind}, a multidimensional interactive attention mechanism developed within the MindSpore framework. Unlike previous approaches that independently model spatial and channel information, MIA-Mind introduces a three-stage pipeline consisting of: (i) global context extraction, (ii) interactive cross-dimensional attention generation, and (iii) dynamic feature reweighting. Specifically, MIA-Mind computes attention maps through cross-multiplicative operations between nonlinear channel descriptors and spatial response maps, enabling more effective context-aware feature fusion with minimal computational overhead. Extensive experiments across diverse tasks—including ResNet-50 on CIFAR-10 for image classification, U-Net on ISBI2012 for medical image segmentation, and CNN-based architectures on CIC-IDS2017 for anomaly detection—validate the generalizability, lightweight design, and performance efficiency of the proposed mechanism.

\section{Methodology}

In this section, we formally define the modeling objective and introduce the structure and implementation details of the proposed Multidimensional Interactive Attention Mechanism (MIA-Mind). The mechanism is designed to enhance task-specific discriminative representations by fusing spatial and channel-wise features through a unified attention scheme. MIA-Mind consists of three functional modules: a global feature extraction module, an interactive attention generation module, and a dynamic reweighting module. The overall architecture is optimized and deployed within the MindSpore framework.

Let the input feature map be denoted as $\mathbf{X} \in \mathbb{R}^{C \times H \times W}$, where $C$, $H$, and $W$ are the number of channels, the height, and the width, respectively. The goal is to learn an attention function $\mathcal{A}(\cdot)$ that outputs a spatial-channel joint attention map $\mathcal{A}_{c,i,j} \in \mathbb{R}^{C \times H \times W}$, used to recalibrate the input features:
\begin{equation}
\mathbf{X}'_{c,i,j} = \mathbf{X}_{c,i,j} \cdot \mathcal{A}_{c,i,j}.
\end{equation}

\begin{figure}[htbp]
    \centering
    \includegraphics[width=0.8\textwidth]{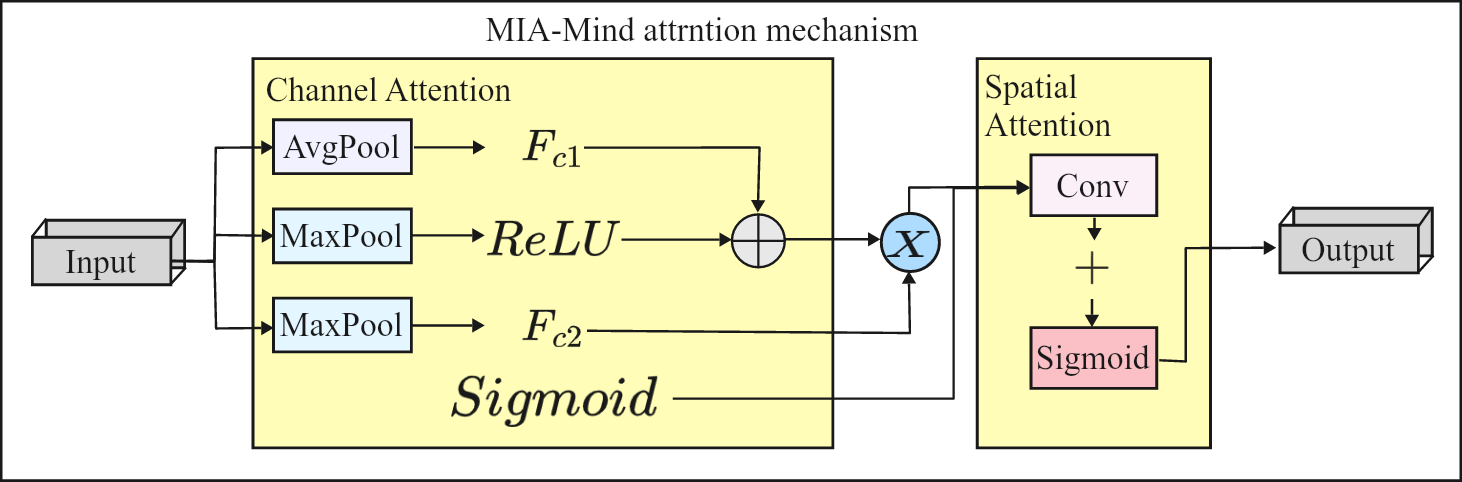}
    \caption{The overall structure of the proposed MIA-Mind attention mechanism, consisting of a channel attention branch and a spatial attention branch that collaboratively recalibrate feature representations.}
    \label{fig:mia_mind_architecture}
\end{figure}

The overall architecture of the proposed MIA-Mind mechanism is illustrated in Figure~\ref{fig:mia_mind_architecture}. MIA-Mind consists of a channel attention branch and a spatial attention branch, which operate in parallel to capture both channel-wise and spatial contextual dependencies. The two branches collaboratively generate multidimensional attention maps that are subsequently used to recalibrate the input features in a lightweight and efficient manner.

\subsection{Global Feature Extraction Module}

The global feature extraction module is responsible for generating initial descriptors that reflect global channel-level and spatial-level context information, serving as the basis for attention computation. Rather than relying on a fixed architecture, MIA-Mind adapts its feature extraction strategy based on the characteristics of the target task. For image classification tasks, we utilize ResNet-50 as the backbone network to obtain residual feature maps with enriched semantic content. For medical image segmentation, where spatial continuity and boundary localization are critical, we integrate UNet to acquire multi-scale hierarchical features. In the case of network traffic anomaly detection, a lightweight CNN model is employed to extract low-dimensional but informative statistical representations of input flows.

Given an intermediate feature map $\mathbf{X} \in \mathbb{R}^{C \times H \times W}$ obtained from one of the aforementioned backbone models, we derive the channel-wise descriptor by applying global average pooling (GAP) over spatial dimensions:
\begin{equation}
\mathbf{z}_c = \frac{1}{H \cdot W} \sum_{i=1}^{H} \sum_{j=1}^{W} \mathbf{X}_{c, i, j}, \quad \text{for } c = 1, \dots, C.
\end{equation}
Simultaneously, a spatial descriptor is generated via channel-wise averaging:
\begin{equation}
\mathbf{M}_{i,j} = \frac{1}{C} \sum_{c=1}^{C} \mathbf{X}_{c, i, j}, \quad \text{for } i = 1, \dots, H; \ j = 1, \dots, W.
\end{equation}

\subsection{Interactive Attention Generation Module}

The goal of this module is to compute a joint attention map that captures the interaction between global channel importance and spatial saliency. Rather than treating channel and spatial dimensions independently, this module establishes a mechanism where channel-wise semantic relevance influences spatial activation patterns, and vice versa.

To encode channel importance, the channel descriptor $\mathbf{z} \in \mathbb{R}^{C}$ is passed through a two-layer fully connected network with a bottleneck architecture:
\begin{equation}
\mathbf{w}_c = \sigma(\mathbf{W}_2 \cdot \delta(\mathbf{W}_1 \cdot \mathbf{z})), \quad \mathbf{w}_c \in \mathbb{R}^{C},
\end{equation}
where $\mathbf{W}_1 \in \mathbb{R}^{\frac{C}{r} \times C}$ and $\mathbf{W}_2 \in \mathbb{R}^{C \times \frac{C}{r}}$ are learnable projection matrices, $r$ is the reduction ratio, $\delta$ denotes ReLU, and $\sigma$ is the sigmoid function.

In parallel, spatial saliency is inferred by applying a convolutional filter over the spatial descriptor $\mathbf{M} \in \mathbb{R}^{H \times W}$:
\begin{equation}
\mathbf{w}_s = \sigma(\mathrm{Conv}_{7 \times 7}(\mathbf{M})), \quad \mathbf{w}_s \in \mathbb{R}^{H \times W}.
\end{equation}

The joint attention map $\mathcal{A} \in \mathbb{R}^{C \times H \times W}$ is derived by combining the two attention representations multiplicatively, allowing the model to account for both per-channel relevance and localized spatial focus:
\begin{equation}
\mathcal{A}_{c,i,j} = \mathbf{w}_c[c] \cdot \mathbf{w}_s[i,j].
\end{equation}

\subsection{Dynamic Reweighting Module}

This module integrates the computed joint attention weights into the original feature map to enhance informative patterns while suppressing irrelevant ones. Each feature point is adaptively reweighted in both channel and spatial dimensions, guided by the previously derived attention scores.

For each element $\mathbf{X}_{c,i,j}$ in the input feature tensor, the recalibrated output is obtained by element-wise multiplication with the corresponding attention coefficient:
\begin{equation}
\mathbf{X}'_{c,i,j} = \mathbf{X}_{c,i,j} \cdot \mathcal{A}_{c,i,j}.
\end{equation}

This unified reweighting strategy ensures that important semantic features are emphasized at appropriate spatial positions, thus improving the model's ability to localize, classify, or detect in task-specific scenarios.

\subsection{MindSpore-based Implementation.} All modules of MIA-Mind are implemented using the MindSpore deep learning framework. Each component is encapsulated in the \texttt{nn.Cell} API for modularity and ease of integration. The attention weights are computed via MindSpore's optimized tensor operations and fused using static computational graphs. During training, we utilize the Adam optimizer with cosine annealing learning rate decay, and enable automatic mixed precision (AMP) and distributed parallel training to ensure efficient deployment on both GPU and Ascend devices.

\section{Experiments}

\subsection{Experimental Setup}

All experiments are conducted using MindSpore 2.5.0 with Python 3.9 on a CPU-only environment. The system configuration includes an Intel Xeon CPU and 64 GB of RAM. No GPU acceleration is used during training or inference phases.
For training, the Adam optimizer is employed with an initial learning rate of 0.01. The batch size is set to 16 across all tasks. Dice loss is adopted as the optimization objective. Each model is trained for 10 epochs without early stopping or additional training tricks.

\subsection{Dataset Description}

To validate the generalization and effectiveness of the proposed MIA-Mind mechanism, experiments are conducted on three publicly available datasets from different domains: natural image classification, biomedical image segmentation, and network anomaly detection.

\textbf{CIFAR-10}: The CIFAR-10 dataset is a widely used benchmark for image classification tasks. It consists of 60,000 color images of size $32 \times 32$ pixels, evenly distributed across 10 distinct classes such as airplane, automobile, bird, cat, and others. The dataset is divided into 50,000 training images and 10,000 test images, with each class containing 6,000 images in total. CIFAR-10 is particularly challenging due to its low resolution and high intra-class variability.

\textbf{ISBI2012}: The ISBI 2012 dataset~\cite{ronneberger2015u} originates from the challenge on segmentation of neuronal structures in electron microscopic (EM) stacks. It contains high-resolution grayscale volumetric images depicting dense cellular structures. The training set comprises a stack of EM slices along with corresponding ground truth labels for membrane segmentation. The dataset poses challenges in terms of fine-grained boundary delineation and the handling of complex tissue structures.

\textbf{CIC-IDS2017}: The CIC-IDS2017 dataset~\cite{sharafaldin2018toward} is a comprehensive benchmark for intrusion detection research. It includes network traffic captures reflecting normal activities as well as a variety of contemporary attack types, such as DDoS, brute force, and botnet activities. The dataset provides a realistic mix of benign and malicious flows, encompassing over 80 extracted network features per flow. Accurate anomaly detection on this dataset is crucial due to the presence of imbalanced and noisy samples.

\subsection{Evaluation Metrics}

To comprehensively assess model performance across different tasks, we adopt a set of standard evaluation metrics. The mathematical definitions of each metric are presented as follows.

\textbf{Accuracy (Acc)} measures the proportion of correctly predicted samples among the total samples:
\begin{equation}
\text{Accuracy} = \frac{TP + TN}{TP + TN + FP + FN},
\end{equation}
where $TP$, $TN$, $FP$, and $FN$ denote true positives, true negatives, false positives, and false negatives, respectively.

\textbf{Precision (Prec)} quantifies the proportion of correctly predicted positive samples among all predicted positives:
\begin{equation}
\text{Precision} = \frac{TP}{TP + FP}.
\end{equation}

\textbf{Recall (Rec)} evaluates the proportion of correctly predicted positive samples among all actual positives:
\begin{equation}
\text{Recall} = \frac{TP}{TP + FN}.
\end{equation}

\textbf{F1-score (F1)} is the harmonic mean of Precision and Recall, balancing their trade-off:
\begin{equation}
\text{F1-score} = 2 \times \frac{\text{Precision} \times \text{Recall}}{\text{Precision} + \text{Recall}}.
\end{equation}

For the segmentation task, we additionally adopt the \textbf{Dice Coefficient (Dice)} to measure the overlap between the predicted segmentation and the ground truth:
\begin{equation}
\text{Dice} = \frac{2 \times |P \cap G|}{|P| + |G|},
\end{equation}
where $P$ and $G$ represent the predicted and ground truth segmentation sets, respectively.

These metrics are reported separately for each task: CIFAR-10 (Accuracy, Precision, Recall, F1-score), ISBI2012 (Accuracy, Dice coefficient), and CIC-IDS2017 (Accuracy, Precision, Recall, F1-score).

\subsection{Experimental Results and Analysis}

The experimental results obtained on the three datasets are summarized in Table~\ref{tab:results}.

\begin{table}[htbp]
\centering
\caption{Performance of MIA-Mind on different tasks and datasets.}
\label{tab:results}
\begin{tabular}{lcccc}
\toprule
\textbf{Task} & \textbf{Dataset} & \textbf{Metric} & \textbf{Score} \\
\midrule
Image Classification & CIFAR-10 & Accuracy & 0.829 \\
                     &          & Precision & 0.831 \\
                     &          & Recall & 0.829 \\
                     &          & F1-score & 0.828 \\
\midrule
Medical Image Segmentation & ISBI2012 & Accuracy & 0.787 \\
                           &          & Dice coefficient & 0.876 \\
\midrule
Anomaly Detection & CIC-IDS2017 & Accuracy & 0.919 \\
                  &             & Precision & 0.989 \\
                  &             & Recall & 0.745 \\
                  &             & F1-score & 0.849 \\
\bottomrule
\end{tabular}
\end{table}

MIA-Mind achieves a classification accuracy of 82.9\% on CIFAR-10, demonstrating its capability to capture discriminative spatial and channel features for natural images. In medical image segmentation, MIA-Mind obtains a Dice coefficient of 87.6\%, indicating excellent boundary localization and fine-grained structure preservation. For network anomaly detection, the model achieves a high overall accuracy of 91.9\%, with outstanding precision but slightly lower recall, suggesting that while most detected anomalies are correct, some attack patterns remain challenging to detect.

\subsection{Experimental Summary}

The experimental results across three representative tasks confirm the effectiveness and versatility of the proposed MIA-Mind mechanism. Without relying on task-specific designs or excessive model complexity, MIA-Mind consistently improves feature representation quality by jointly modeling spatial and channel interactions. 

Through systematic evaluations on classification (CIFAR-10), segmentation (ISBI2012), and anomaly detection (CIC-IDS2017) tasks, the experiments demonstrate that MIA-Mind achieves significant performance gains in terms of accuracy, boundary localization, and anomaly sensitivity. These findings highlight the general applicability of the proposed method across heterogeneous domains, validating its modular design philosophy and computational efficiency under MindSpore's lightweight framework.

The experiments further demonstrate that even under CPU-only environments and relatively limited computational resources, MIA-Mind can achieve competitive results, illustrating the practicality of the design for real-world deployment scenarios where hardware resources may be constrained.

\section{Discussion}

The experimental results demonstrate that MIA-Mind consistently achieves strong performance across classification, segmentation, and anomaly detection tasks. This can be attributed not only to the proposed multidimensional interactive attention mechanism but also to the efficient computational support provided by the MindSpore framework. MindSpore’s optimized graph compilation and operator fusion strategies significantly reduce the runtime latency and memory footprint, allowing the model to operate efficiently even under CPU-only settings. The modular Cell API in MindSpore further facilitates seamless integration of the MIA-Mind module into various backbone architectures, maintaining both flexibility and computational efficiency.

MIA-Mind exhibits several notable advantages compared to conventional attention mechanisms. Firstly, the multidimensional interactive structure enables simultaneous modeling of spatial saliency and channel importance, capturing complex feature dependencies that are often neglected by separable attention designs. Secondly, the module maintains a lightweight computational profile by employing bottleneck transformations and element-wise fusion strategies, avoiding significant overhead compared to standard convolutional blocks. Thirdly, the modular design of MIA-Mind ensures high compatibility and plug-and-play capability with diverse neural network architectures, enhancing its applicability across multiple tasks without necessitating substantial architectural modifications.

Despite the promising results, the current study has several limitations. First, the experiments are conducted primarily under single-device, CPU-only settings, which may not fully reflect the scalability of MIA-Mind in large-scale, distributed training environments. Second, while the mechanism shows strong generalization across three domains, further validation on larger and more complex datasets, such as ImageNet-1K or Cityscapes, is necessary to assess its robustness. Third, the current attention fusion strategy is static; exploring dynamic or data-adaptive fusion mechanisms could further enhance model adaptability and performance. Future research will focus on addressing these issues by extending MIA-Mind to distributed GPU clusters, applying it to broader real-world datasets, and introducing adaptive attention fusion techniques.

The integration of MIA-Mind within the MindSpore framework highlights the practical advantages of this combination. MindSpore’s automatic parallelism, static graph optimization, and lightweight operator support enable MIA-Mind to achieve competitive results even in resource-constrained environments. This synergy not only validates the computational efficiency of MindSpore for attention-intensive models but also showcases MIA-Mind’s potential for deployment in real-world applications where hardware limitations and inference efficiency are critical considerations.

\section{Conclusion}

In this study, we proposed MIA-Mind, a multidimensional interactive attention mechanism designed to enhance feature representation by jointly modeling channel importance and spatial saliency. Built upon the MindSpore framework, MIA-Mind effectively integrates lightweight computation with flexible modularity, enabling seamless incorporation into diverse neural network architectures.

Extensive experiments conducted on CIFAR-10, ISBI2012, and CIC-IDS2017 datasets demonstrate that MIA-Mind consistently improves classification accuracy, segmentation boundary delineation, and anomaly detection sensitivity. Notably, the mechanism achieves competitive performance under CPU-only settings, highlighting its efficiency and practical deployability.

The contributions of this work are three-fold: (i) introducing a unified multidimensional attention fusion strategy that captures cross-dimensional feature interactions; (ii) achieving task-general improvements without introducing significant computational overhead; and (iii) validating the design within the efficient MindSpore ecosystem, showcasing its real-world applicability.

Future research will focus on extending MIA-Mind to larger-scale datasets, exploring adaptive attention fusion strategies, and investigating distributed training implementations to further enhance scalability and robustness.

\section*{Acknowledgments}
Thanks for the support provided by the MindSpore Community.

\vspace{6pt}

\bibliographystyle{unsrt}
\bibliography{references}

\end{document}